\documentclass[10pt,journal,compsoc]{IEEEtran}
\ifCLASSOPTIONcompsoc
  \usepackage[nocompress]{cite}
\else
  \usepackage{cite}
\fi

\usepackage{amsmath}
\usepackage{amsfonts,color}
\usepackage{mathrsfs}
\usepackage{latexsym,amssymb}
\usepackage{graphicx}
\usepackage{cite}
\usepackage{subfigure,balance}
\usepackage{fancyhdr}  %add footnote
\usepackage{graphicx}
\usepackage{algorithmic}
\usepackage{graphicx}
\usepackage{textcomp}
\usepackage{hyperref}
\usepackage{stfloats}
\usepackage{float}
\usepackage{amsmath}
\usepackage{amsfonts,color}
\usepackage{mathrsfs}
\usepackage{latexsym,amssymb}
\usepackage{graphicx}
\usepackage{cite}
\usepackage{subfigure,balance}
\usepackage{fancyhdr}
\usepackage{makecell}
\usepackage{booktabs}
\usepackage{algorithmic}
\usepackage{booktabs}
\usepackage{ragged2e}
\usepackage{tabularx}
\usepackage{bbding}
\usepackage{pifont}
\usepackage{wasysym}
\usepackage{amssymb}
\usepackage{caption}
\usepackage{multirow} % Required for multirows
\usepackage{enumitem,amssymb}
\newlist{todolist}{itemize}{2}
\setlist[todolist]{label=$\square$}
\usepackage{pifont}
\ifCLASSINFOpdf
\else
\fi
\begin{document}
\title{A Multimodal Data-driven Framework for Anxiety Screening}

\author{Haimiao~Mo,
         Shuai Ding*,~\IEEEmembership{Member, IEEE,}
        Siu Cheung Hui % <-this % stops a space
\IEEEcompsocitemizethanks{
\IEEEcompsocthanksitem 
\textbf{Haimiao Mo} and \textbf{Shuai Ding} are with the School of Management, Hefei University of Technology, Anhui Hefei 23009, China, also with the Key Laboratory of Process Optimization and Intelligent Decision-Making, Ministry of Education, China. (Email: mhm\_hfut@163.com, dingshuai@hfut.edu.cn)\protect\\

\IEEEcompsocthanksitem 
\textbf{Siu Cheung Hui} is with the  School of Computer Science and Engineering, Nanyang Technological University, Singapore 639798. (Email: ASSCHUI@ntu.edu.sg)
 } }

\IEEEtitleabstractindextext{%
\begin{abstract}
\justifying 
Early screening for anxiety and appropriate interventions are essential to reduce the incidence of self-harm and suicide in patients. Due to limited medical resources, traditional methods that overly rely on physician expertise and specialized equipment cannot simultaneously meet the needs for high accuracy and model interpretability. Multimodal data can provide more objective evidence for anxiety screening to improve the accuracy of models. The large amount of noise in multimodal data and the unbalanced nature of the data make the model prone to overfitting. However, it is a non-differentiable problem when high-dimensional and multimodal feature combinations are used as model inputs and incorporated into model training. This causes existing anxiety screening methods based on machine learning and deep learning to be inapplicable. Therefore, we propose a multimodal data-driven anxiety screening framework, namely MMD-AS, and conduct experiments on the collected health data of over 200 seafarers by smartphones. The proposed framework's feature extraction, dimension reduction, feature selection, and anxiety inference are jointly trained to improve the model's performance. In the feature selection step, a feature selection method based on the Improved Fireworks Algorithm is used to solve the non-differentiable problem of feature combination to remove redundant features and search for the ideal feature subset. The experimental results show that our framework outperforms the comparison methods.

\end{abstract}

}

\maketitle

\IEEEdisplaynontitleabstractindextext

% \fi
\IEEEpeerreviewmaketitle

\IEEEraisesectionheading{\section{Introduction}\label{sec:introduction}}


\IEEEPARstart{I}{n} 2019, mental illnesses, particularly anxiety disorders, were not only among the top twenty-five leading causes of excess global health spending, but also among the most disabling mental illnesses \cite{gbd2022global}. Furthermore, anxiety disorders are accompanied by immune disorders \cite{hou2016abstract}, and interfere with cognitive functions through memory and attention \cite{dias2009chronic}, thereby affecting normal life and work. Early anxiety assessment and appropriate interventions can greatly reduce the rate of self-harm and suicide in patients \cite{enock2014attention}.

Psychological scales and routine health checks with professional medical equipment are traditional anxiety screening methods. The Self-rating Anxiety Scale (SAS) \cite{baygi2021prevalence}  and the Generalized Anxiety Disorder-7 (GAD-7) \cite{stocker2021patient} are two psychological scales that are currently used for anxiety screening. Anxiety frequently results in a variety of symptoms or behavioral modifications, such as breathlessness \cite{efinger2019distraction}, variations in blood pressure \cite{pham2021heart} and heart rate \cite{petrescu2020integrating}, perspiration, tense muscles, and dizziness  \cite{giannakakis2017stress}. These objective signs can also be used as an important basis for anxiety screening. However, due to the limitation of lacking of medical resources in remote areas and high cost, routine health examinations such as Magnetic Resonance Imaging (MRI) \cite{pfurtscheller2021mri}, Computed Tomography (CT), electrocardiogram (ECG) \cite{ihmig2020line}, \cite{puli2019toward } and electroencephalogram (EEG) \cite{petrescu2020integrating}, \cite{giannakakis2015detection}, may not be available.

Noncontact screening methods are another typical anxiety screening tool. They usually use computer vision or deep learning techniques to extract the behavioral or physiological features for anxiety screening. These methods have the advantages of low cost and convenience. The application of behavioral features \cite{giannakakis2017stress}, \cite{zhao2019see}, speech features and text features provides more objective evidence for anxiety screening. Moreover, physiological signals, such as heart rate \cite{giannakakis2017stress}, \cite{petrescu2020integrating}, heart rate variability \cite{ihmig2020line}, and respiration rate, can be obtained by imaging photoplethysmography (iPPG) technology  \cite{favilla2018heart}, which can also be used as important features for anxiety screening.

Due to the complicated genesis and protracted nature of mental diseases\cite{farre2017new}, diagnosing them frequently involves knowledge from a number of fields, including biomedicine, psychology, and social medicine. It is a challenging problem to obtain more timely multimodal information about patient's health using traditional medical screening methods due to the limitation of medical resources \cite{baygi2021prevalence}. In addition, multimodal data can also provide more objective evidence \cite{lee2020detection} to improve the accuracy of anxiety screening. Therefore, multimodal data will be the driving force behind the future development of anxiety screening\cite{zhang2020fusing}.

However, the large amount of noise in the multimodal data and the imbalance of the data make the model prone to overfitting. In other words, the model cannot screen anxious patients with high precision and take intervention measures in advance, which may have a negative impact on their lives or mental conditions. In addition, due to the poor medical conditions in remote areas, model interpretability \cite{stiglic2020interpretability} and important features are crucial to assist primary care staff in anxiety screening. Traditional machine learning methods \cite{puli2019toward}, \cite{ancillon2022machine} are difficult to deal with the problem of scalability and generalization of multimedia content data in a fast and accurate way. Deep learning methods \cite{lee2020detection}, \cite{vsalkevicius2019anxiety} based on computer vision have higher robustness and accuracy compared with traditional methods, and are therefore increasingly widely used for anxiety screening. Most of the existing technologies for anxiety screening focus on differentiable optimization problems. And the combination of high-dimensional and multimodal features is a non-differentiable problem when used as input to the model and incorporated into the model training. These existing methods are unable to meet the requirement on both the high accuracy and model-interpretability scenarios for anxiety screening. Therefore, we propose a Multimodal Data-driven framework for Anxiety Screening (MMD-AS). 

The contributions of this paper are as follows.

\begin{itemize}
	
	\item 
	We propose a low-cost, noncontact, interpretable and easy-to-use anxiety screening framework that enables multimodal data capture and remote anxiety screening via smartphones only, which is suitable for scenarios with limited medical resources, such as health protection for seafarers on long voyages and mental health screening in remote areas.
	
	\item 
	To improve the performance, the framework's components are jointly trained. In addition, our Improved Fireworks Algorithm (IFA) solves the non-differentiable problem in the case of feature combination by enhancing the local search capability, which filters out redundant features and reduces the noise in the data to find the best feature subset.

	\item 
	Experimental results of anxiety screening in more than 200 seafarers show that our framework has achieved high precision and model interpretability. More importantly, the results point out that multimodal data is essential for anxiety screening, and the important indicators for anxiety detection are identified, which are both beneficial to clinical practice.
	
\end{itemize}

The rest of this paper is organized as follows. Section 2 reviews the related work on anxiety representations, anxiety screening, feature extraction and multimodal data-driven methods. Section 3 presents our proposed framework for anxiety screening. Section 4 presents the performance evaluation. Sections 5 discusses the limitations of the proposed framework. Finally, Section 6 concludes the paper.

\section{RELATED WORK}
In this section, we review the related work on anxiety representations, anxiety screening, feature extraction and multimodal data-driven methods. Table 1 summarizes the features and methods for anxiety screening.

\subsection{Anxiety Representation}

Anxiety is a feeling of tension, worry, or restlessness. It occurs frequently in a variety of mental conditions, including phobias, panic disorders, and generalized anxiety disorders \cite{bandelow2022epidemiology}. Anxiety is a typical response to risk and mental stress. The amygdala and hippocampus are activated by the feelings of fear and dread brought on by stress, which also affects the autonomic and parasympathetic nervous systems \cite{shin2010neurocircuitry}. Patients with anxiety disorders exhibit physical symptoms that are linked to the disease, such as rapid breathing \cite{efinger2019distraction}, heartbeat, BP \cite{pham2021heart}, and additional symptoms \cite{gavrilescu2019predicting} such as perspiration, tightness in the muscles, and vertigo. The physiological signals that are most frequently utilized in physiological research to evaluate mental health include ECG \cite{elgendi2019assessing}, heart rate, heart rate variability \cite{pham2021heart}, EEG \cite{li2019recognition}, and electrode signals, as shown in Table 1.

Patients with anxiety disorders exhibit structural and functional abnormalities in the nerve system that regulates emotion, according to brain imaging studies. As shown in Table 1, a person's ability to manage their emotions can be plainly noticed in their facial and behavioral characteristics \cite{giannakakis2017stress} as well as audio indicators (such as intonation and speech tempo) \cite{ozseven2018voice}. For example, the insula, frontal orbital cortex, anterior cingulate cortex, striatum, and amygdala all exhibit diminished response to unfavorable emotional stimuli \cite{martin2010neurobiology}. Due to physiological respiratory issues, individuals with anxiety disorders have voices that are a mirror of their circumstances. Related studies have shown that anxious patients exhibit elevated wavelet, jitter, shimmer, and fundamental frequency (F0) mean coefficients \cite{albuquerque2021association}. Mel-Frequency Cepstral Coefficients (MFCCs) declines in the presence of anxiety \cite{ozseven2018voice}. The main signs of facial anxiety include changes in the eyes, including pupil size variations, blink rates \cite{zhang2020fusing}, and gaze distribution \cite{mogg2007anxiety}, as well as in the lips, including lip twisting and mouth movement. Other key facial anxiety indicators include changes in the cheeks, head movement, and head speed \cite{adams2015decoupling}. Additional facial signs of anxiety include pallor, twitching eyelids, and stiffness in the face. Numerous studies have shown a link between the size of the pupil and emotional or mental activity. Dilated pupils may be a sign of higher anxiety levels \cite{bradley2008pupil}. The coherence and direction of the eyes, are also impacted by anxiety \cite{mogg2007anxiety}. Increased gaze volatility during voluntary and stimulus-driven gazing are correlated with high levels of trait anxiety \cite{guo2019change}. For instance, those who are nervous typically scan negative stuff more than those who are not anxious \cite{giannakakis2017stress}.

\begin{table*}[!htbp]
	\centering
	\caption{Features, feature extraction, feature selection and classification methods for anxiety screening.}
	\label{TABLE1}
	\scalebox{0.98}{
		\begin{tabularx}{\textwidth}{p{7.2cm} p{3.3cm}p{3.1cm}p{3.1cm}}
			\toprule
			\textbf{Features} & \textbf{Feature extraction }&\textbf{Feature selection} &\textbf{Classification} \\
			\midrule
			\textbf{\textit{1)} Physiological features: }
			
			Heart Rate (HR)  
			\cite{giannakakis2017stress}, \cite{petrescu2020integrating},
			\cite{ihmig2020line},
			Heart Rate Variability (HRV) 
			\cite{ihmig2020line},
			Respiratory Rate (RR) \cite{haritha2017automating},
			Blood Pressure (BP) 
			\cite{pham2021heart},
			electroencephalogram (EEG) 
			\cite{petrescu2020integrating}, \cite{giannakakis2015detection},
			\cite{zhang2020fusing},
			\cite{lee2020detection},
			electrocardiography (ECG) 
			\cite{ihmig2020line}, 
			\cite{puli2019toward}, \cite{li2019recognition},
			Electrodermal Activity (EDA) 
			\cite{giannakakis2017stress},  \cite{petrescu2020integrating},
			\cite{ihmig2020line},
			and imaging photoplethysmography (iPPG)  
			\cite{giannakakis2017stress}, \cite{lee2020detection}.
			
			\textbf{\textit{2)} Behavioral features:}
			
			\textit{Eyes}: 
			gaze spatial distribution and gaze direction \cite{giannakakis2017stress},
			saccadic eye movements \cite{giannakakis2017stress},
			\cite{richards2011influence}, \cite{zhang2020fusing}, \cite{lisk2020systematic},
			pupil size \cite{giannakakis2017stress} and pupil ratio variation \cite{giannakakis2017stress},
			blink rate, eyelid response, eye aperture, eyebrow movements \cite{giannakakis2017stress};
			\textit{Lip}: lip deformation, lip corner puller/depressor and lip pressor \cite{giannakakis2017stress};
			\textit{Head}:
			head movement \cite{giannakakis2017stress};
			\textit{Mouth}: mouth shape \cite{giannakakis2017stress}; \textit{Gait} \cite{zhao2019see};
			\textit{Motion data} \cite{puli2019toward}.
			
			\textbf{\textit{3)} Audio features:}
			
			Wavelet, jitter, shimmer, F0 mean coefficients \cite{albuquerque2021association} and Mel-frequency cepstral coefficients (MFCCs) \cite{ozseven2018voice}.
			
			\textbf{\textit{4)} Text features:}
			
			Demographics \cite{sau2019screening}, occupation and health;
			blog posts \cite{tyshchenko2018depression},
			semantic location \cite{saeb2017mobile}.
			
			\textbf{\textit{5)} Questionnaires features:} SAS \cite{baygi2021prevalence} and GAD-7 items \cite{stocker2021patient}.
			& 
			\textbf{\textit{1)} Neural networks (NN):}
			
			Convolutional Neural Network (CNN)
			\cite{petrescu2020integrating},
			\cite{tyshchenko2018depression},
			\cite{richards2011influence},
			\cite{umrani2022hybrid},
			Long Short-Term Memory (LSTM)
			\cite{richards2011influence},
			Radial Basis Function (RBF)
			\cite{vsalkevicius2019anxiety}, Artificial Neural Networks (ANN) \cite{umrani2022hybrid}, \cite{lee2020detection},
			and Generalized Likelihood Ratio (GLR) \cite{giannakakis2017stress}.
			
			\textbf{\textit{2)} Correlation analysis:}
			
			Principal Component Analysis (PCA) \cite{haritha2017automating}, 
			Canonical Correlation Analysis (CCA), Sparse Canonical Correlation Analysis (SCCA) \cite{zhang2020fusing}.
			
			\textbf{\textit{3)} Signal processing method:}
			
			Kalman filter \cite{puli2019toward}, fast Fourier transform \cite{mo2022collaborative}.
			&
			\textbf{\textit{1) }Filter methods:}
			
			informative properties-based  includes Random Forest (RF) with Gini index \cite{vsalkevicius2019anxiety};
			saliency tests-based, such as 
			SKB,  correlation analysis \cite{petrescu2020integrating},
			\cite{zhang2020fusing}, \cite{doty2013fearful}, Pearson Correlation Coefficient (PCC) 
			\cite{zhao2019see}, \cite{richards2011influence}, and \textit{t}-test \cite{richards2011influence}.

			\textbf{\textit{2)} Wrapper methods:} 
			
			sequential feature selection \cite{richards2011influence}, 
			such as Sequential Backward Selection (SBS) and Sequential Forward Selection (SFS) methods\cite{giannakakis2015detection};
			iterative-based includes RFE \cite{sau2019screening}.
			& 
			\textbf{\textit{1)} Machine learning methods:}
			
			Support Vector Machines (SVM) 
			\cite{giannakakis2017stress},
			\cite{ihmig2020line},
			\cite{puli2019toward}, 
			\cite{tyshchenko2018depression},
			\cite{richards2011influence}, \cite{vsalkevicius2019anxiety}, \cite{sau2019screening},   
			\cite{zhang2020fusing},
			Logistic Regression (LR) 
			\cite{giannakakis2017stress},
			\cite{puli2019toward}, 
			\cite{zhao2019see},
			\cite{sau2019screening},
			\cite{umrani2022hybrid}, \cite{li2019recognition},
			DTs
			\cite{ihmig2020line},
			\cite{puli2019toward},
			RF
			\cite{tyshchenko2018depression},
			\cite{richards2011influence},
			\cite{sau2019screening}, \cite{umrani2022hybrid},
			Naïve Bayes (NB)
			\cite{giannakakis2017stress}, \cite{ihmig2020line},
			\cite{sau2019screening},
			K-Nearest Neighbors (KNN)
			\cite{giannakakis2017stress}, 
			\cite{puli2019toward},
			\cite{umrani2022hybrid},
			\cite{zhang2020fusing}, \cite{li2019recognition},
			Adaptive Boosting (AdaBoost)
			\cite{giannakakis2017stress}, \cite{puli2019toward},
			XGBoost 
			\cite{saeb2017mobile},
			Catboost 
			\cite{puli2019toward},
			\cite{sau2019screening}.
			
			\textbf{\textit{2)} methods based on NN: }
			
			CNN, LSTM, RBF, GLR, ANN.
			\\\midrule
		\end{tabularx}%
	}
\end{table*}%

\subsection{Anxiety Screening Methods}

Traditional and noncontact screening methods are based on machine learning or deep learning techniques, which are the two primary categories of anxiety screening techniques. Psychological scales, \cite{stocker2021patient} and assisting technologies (such as  MRI \cite{pfurtscheller2021mri}, CT, ECG \cite{elgendi2019assessing}, and EEG \cite{li2019recognition}, and biochemical indicators \cite{hou2017peripheral}) are frequently used in traditional screening approaches to evaluate anxiety levels. Noncontact screening methods mainly use computer vision  \cite{gavrilescu2019predicting} or deep learning techniques \cite{puli2019toward}, \cite{mo2022collaborative} to extract behavioral characteristics and physiological signals related to anxiety.

\subsubsection{Traditional Screening Methods}

Traditional mental health examinations commonly use psychological scales, such as the SAS \cite{baygi2021prevalence} and GAD-7 \cite{stocker2021patient} to ascertain whether patients are suffering anxiety. In real clinical settings, doctors routinely conduct structured interviews with patients to find out more about their mental health. The patient's body language and facial emotions should be closely observed by the doctor all the time. This method is severely constrained by the interactions between the doctor and patient as well as by the expertise and experience of psychiatrists. To make the proper diagnosis, medical professionals may also take into account additional data from tests such as MRI \cite{pfurtscheller2021mri}, CT, ECG \cite{elgendi2019assessing}, and EEG \cite{li2019recognition}. To find people who might have psychiatric problems, extensive biological data gathering is also carried out on them, such as monitoring inflammatory markers and hormone changes \cite{hou2017peripheral}. However, traditional screening methods place an undue emphasis on psychiatrists' training and experience. Traditional detection methods fall short in the face of unique situations, such as on long-distance voyages with limited medical resources \cite{tang2022seafarers}.

\subsubsection{Noncontact Screening Methods}

Changes in behavioral characteristics, such as concentration on things (reflected in eye gaze duration \cite{mogg2007anxiety}, eye movement characteristics \cite{guo2019change}, pupil size, and changes in head posture  \cite{baltrusaitis2018openface}), mouth shape, eyebrow shape \cite{giannakakis2017stress}, facial expression \cite{sani2018mood}, and gait \cite{zhao2019see}, can reflect some extent the person's mental activity. These mental activities can lead to significant changes in a person's physiological characteristics, such as EEG \cite{li2019recognition}, heart rate \cite{ihmig2020line}, and respiration rate. Noncontact anxiety screening methods capture or extract these behavioral or EEG changes mainly through computer vision or signal processing methods. A facial action coding system \cite{gavrilescu2019predicting} is often used to characterize a person's facial behavior. To explore the relationship between behavioral and physiological features and anxiety, Canonical Correlation Analysis (CCA) methods \cite{petrescu2020integrating} such as the Pearson Correlation Coefficient (PCC) \cite{zhao2019see}, \cite{richards2011influence} are commonly used. Moreover, because of the physiological-behavioral link, machine learning methods can even perform better by incorporating EEG and eye-movement features. However, since correlation analysis methods tend to cause overfitting of classical machine learning methods such as SVM and KNN, sparse representation methods address this problem by introducing constraint terms \cite{zhang2020fusing}. To reduce the risk of model overfitting, sequence-based feature selection approaches \cite{giannakakis2015detection} such as Sequential Backward Selection (SBS) and Sequential Forward Selection (SFS) are used to remove redundant features from the original data.

The link between physiological symptoms and mental illness has led researchers to focus on additional identification of mental illness through physiological characteristics. Elevated heart rate, rapid breathing \cite{pfurtscheller2022processing}, high BP \cite{pham2021heart}, dizziness, sweat, and muscle tension can all be used to objectively screen for anxiety. However, specialized hardware is needed for the traditional gathering of physiological signals. Its relatively high cost frequently prevents it from providing early diagnosis of psychiatric diseases.

Physiological characteristics \cite{ding2022noncontact} such as blood volume pulse, heart rate, heart rate variability, and respiratory rate can be captured by signals from imaging photoplethysmography (iPPG). In fact, the iPPG signals are extracted by computer vision technology. Affordableness, noncontact, safety, the ability to obtain continuous measurements, and ease of use are just a few advantages of iPPG \cite{favilla2018heart}. In the research of noncontact telemedicine monitoring and physical and mental health monitoring, iPPG offers a novel perspective.

\subsection{Feature Extraction Methods}

As shown in Table 1, the feature extraction methods for anxiety screening are mainly classified into three categories: neural network-based, correlation analysis, and signal processing methods. High data dimensionality and data redundancy are properties of the time series data (such as ECG \cite{zhang2020fusing} and EEG \cite{li2019recognition}) and image data (such as CT,  MRI \cite{pfurtscheller2021mri}) utilized for anxiety inference. The performance of screening anxiety methods may thus be adversely affected. To improve the performance of the traditional screening methods, the feature extraction methods in Table 1, such as Neural Networks (NN) \cite{petrescu2020integrating}, \cite{tyshchenko2018depression}, \cite{vsalkevicius2019anxiety} and correlation analysis \cite{zhao2019see}, \cite{richards2011influence}, are used to extract features useful for anxiety inference. Nonlinear information useful for anxiety inference is frequently extracted using neural network-based feature extraction techniques including CNN, LSTM, and Radial Basis Function (RBF). 

Canonical Correlation Analysis (CCA), Sparse Canonical Correlation Analysis (SCCA) \cite{zhang2020fusing}, and Principal Component Analysis (PCA) \cite{haritha2017automating} are correlation analysis techniques for the extraction of anxiety-related features, which are helpful for inferring anxiety. In \cite{haritha2017automating}, PCA uses orthogonal transformations to project observations of a group of possibly correlated variables (such as the time and frequency domain statistical from respiratory signals) into the principle components of a group of linearly uncorrelated variables. The number of features in a dataset can be controlled by imposing a threshold using PCA, which is frequently used to minimize the dataset's dimensions.

Due to acquisition equipment or environmental factors, the quality of physiological signal extraction or analysis is easily affected. For example, the iPPG signals used for anxiety inference is particularly susceptible to ambient lighting and motion artifacts \cite{mo2022collaborative}. So these physiological features from the iPPG signals contain a lot of noise. The signal processing techniques in Table 1, such as  Kalman filter \cite{puli2019toward} and fast Fourier transform \cite{mo2022collaborative}, are typically used to reduce noise and  eliminate its detrimental effects on the anxiety screening model. In \cite{puli2019toward}, the enhanced Kalman filter processes the heart rate and accelerometry signals to follow the user's heart rate information in various contexts and choose the best model for anxiety detection based on the user's exercise conditions.

\subsection{Multimodal Data-Driven Methods}

Due to complex etiology and long development cycle, the diagnosis of mental illness usually requires a multidisciplinary approach that combines biomedicine, psychology, and social medicine \cite{farre2017new}. Moreover, multimodal data can provide more objective evidence for anxiety screening \cite{lee2020detection}. The multimodal data-driven approaches are promising for anxiety screening. By using the correlation analysis approach to examine the structural information between the EEG and eye movement data, it is possible to combine these two types of variables and detect anxiety more precisely \cite{zhang2020fusing}. Biophysical signals such as heart rate, skin electricity, and EEG in virtual reality applications, are extracted as features of different dimensions \cite{petrescu2020integrating}. The features from the time domain and frequency domain are then fused to achieve high-precision anxiety detection. Several biosignals, including EEG, iPPG, Electrodermal Activity (EDA), and pupil size \cite{lee2020detection}, are used to measure anxiety in various driving scenarios.

However, the use of multimodal data will necessarily result in a rapid increase in data dimensions and feature redundancy. The increase in data dimensions leads to curse of dimensionality. The model's accuracy may be affected since redundant features may have a lot of noise. Decision Trees-based (DTs) approaches such as Random Forest (RF), Adaptive Boosting (AdaBoost) \cite{giannakakis2017stress}, Naïve Bayes (NB) \cite{sau2019screening}, extreme gradient boost (XGBoost) \cite{saeb2017mobile}, can select features useful for anxiety detection. They use information attributes (such as information gain, information entropy, or Gini index \cite{vsalkevicius2019anxiety}) to lessen the dimensionality and number of redundant features in the data. Besides, unbalanced datasets are very common, particularly in the medical industry. It also poses a significant obstacle to anxiety screening. The existing machine learning techniques  \cite{giannakakis2017stress}, \cite{puli2019toward}, \cite{tyshchenko2018depression},  \cite{sau2019screening}, \cite{ancillon2022machine} and neural network techniques  \cite{giannakakis2017stress}, \cite{tyshchenko2018depression}, \cite{richards2011influence}, \cite{vsalkevicius2019anxiety}, \cite{lee2020detection} used for anxiety screening are therefore challenging to apply to this unique situation.

\section{PROPOSED FRAMEWORK}

\begin{figure*}[!ht]
	\begin{center}
		\includegraphics[width = 15cm]{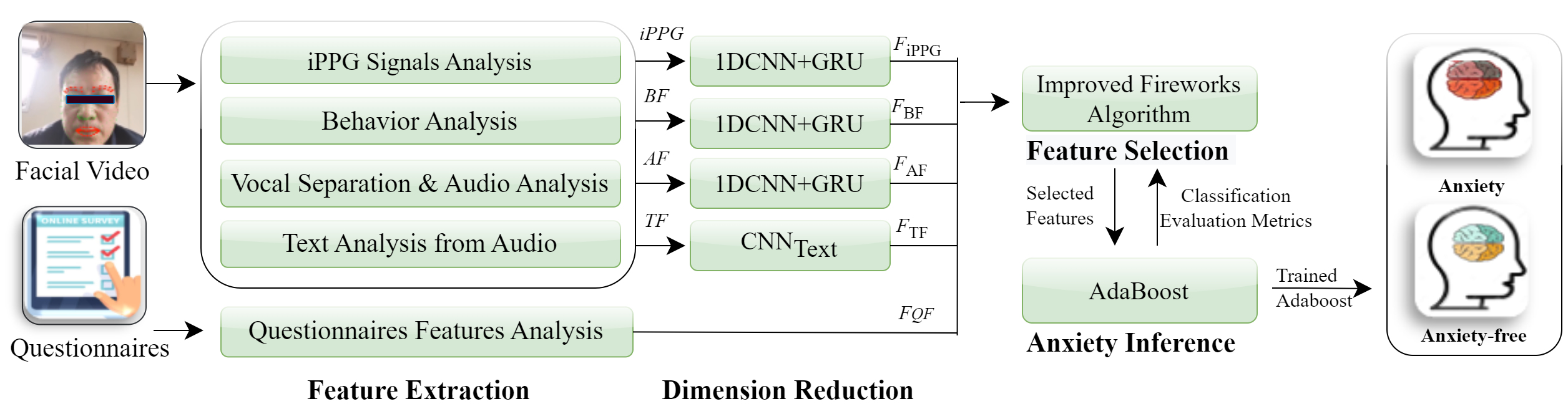}
	\end{center}
	\caption{The framework for anxiety screening.}
	\label{Fig0}
\end{figure*}

Figure 1 shows the proposed Multimodal Data-driven framework for Anxiety Screening (MMD-AS) which consists of four main components: feature extraction, dimension reduction, feature selection, and anxiety inference for seafarers. The steps of the framework are as follows. First, feature extraction extracts the heart rate and respiratory rate features (iPPG), Behavior Features (BF), and Audio Features (AF) from facial videos. The Text Features (TF) are extracted from audio. The Questionnaires Features (QF) are extracted by processing the questionnaire. Next, the extracted iPPG, BF, AF, and TF features are processed by the designed "1DCNN+GRU" and "CNN$_{Text}$" networks for dimension reduction, and then combined with the QF feature to create the feature vector $F = [F_{iPPG}, F_{BF}, F_{AF}, F_{TF}, F_{QF}]$. Then, feature selection selects the feature subsets from the vector $F$ based on the Improved Fireworks Algorithm (IFA). Finally, the data from the selected features are trained by the AdaBoost classifier for evaluating the generalization ability of the selected features based on the classification evaluation metrics. The trained model and selected features are finally used for anxiety inference.

\subsection{Feature Extraction} 
Table 2 shows the different features extracted in Feature Extraction. The heart rate and respiration rate features from iPPG signals, BF, AF, TF, and QF features from facial video and questionnaires, which are used for anxiety inference.

\subsubsection{ Heart Rate and Respiration Rate Features}

\begin{table*}[htbp]
	\centering
	\caption{Features extracted from facial video and questionnaires.}
	\label{TABLE2}
	\scalebox{0.98}{
		\begin{tabularx}{\textwidth}{p{6.6cm} p{11cm}}
			\toprule
			Feature types  & Description\\ \midrule
			\textbf{\textit{1)} Heart Rate and Respiration Rate Features:}
			\begin{itemize}
				\item iPPG${_{Nose}^{TD}}$,  iPPG${_{Nose}^{TD}}$, iPPG${_{Fore}^{FD}}$, and iPPG${_{Fore}^{FD}}$ 
			\end{itemize}
			& The iPPG signals extracted from the nose and forehead regions containing time and frequency domain features of heart rate (HR) and respiration rate (RR). \\
			\toprule
			
			\textbf{\textbf{\textit{2)} Behavioral Features (BF):}} 
			\begin{itemize}
				\item Head Position (HP)
				\item Eye Gaze (EG)
				\item Action Unit (AU): AU01, AU02, AU04, AU05, AU06, AU07, AU09,  AU10, AU12, AU14,  AU15, AU17, AU20, AU23, AU25, AU26, AU44,  AU45
			\end{itemize}
			& 
			1) HP=($x_{HP}$, $y_{HP}$, $z_{HP}$) describes the positional information of the head rotation when the center of the head is taken as the origin.
			
			2) EG=($x_{EG}$, $y_{EG}$) describes the eye gaze direction.
			
			3) AUs are defined by the facial action coding system \cite{liu2019facial}, which refers to the set of facial muscle movements corresponding to the displayed emotions.\\
			\toprule
			
			\textbf{\textit{3)} Audio Features (AF): }
			\begin{itemize}
				\item Mel-frequency cepstral coefficients (MFCCs), fundamental frequency F0, Zero-Crossing Rate (ZCR), 
				Prosody, 
				and Phonation
			\end{itemize}
			& 
			The audio in the video uses the vocal separation method to obtain audio from human voices. Then, AF are extracted by audio analysis technology \cite{pariente2020asteroid}.
			Phonation features include F0's first and second derivatives  ($F0^1$ and $F0^2$), jitter, shimmer, Amplitude Perturbation Quotient (APQ), Pitch Perturbation Quotient (PPQ), and Logarithmic Energy (LE).\\
			\toprule
			
			\textbf{\textit{4)} Text Features (TF): } 
			\begin{itemize}
				\item Text features from audio
			\end{itemize}
			&
			Process the text in the audio through the iFlytek toolkit \cite{liu2022spontaneous}, and then extract the text features using the pre-trained  BERT model \cite{devlin2018bert}.

			\\\toprule
			
			\textbf{\textit{5)} Questionnaires Features (QF):}
			\begin{itemize}
				\item Assessment Time (AT)
				\item Personal Information (PF)
				\item Big Five Personalities Traits (BFPT) \cite{nikvcevic2021modelling} 
				\item Sleep Quality (SQ) \cite{mollayeva2016pittsburgh}
				\item Lifestyle
				\item Emotional State (ES)
				\item Work Environment (WE) \item Entertainment (En)
				\item Attitude to Life (AL) 
				\item Social Support (SS) 
				\item Family Relationships (FR) 
			\end{itemize}
			&
			1) AT: It includes the stages of before boarding, sailing, and after disembarking.

			2) PF: It includes marital status, family size, income, place of household registration, position, working hours, smoking and alcohol use \cite{sau2019screening}.
			
			3) BFPT: It includes extraversion, agreeableness, openness, conscientiousness, and neuroticism. 
			
			4) SQ: It is evaluated by Pittsburgh Sleep Quality Index (PSQI) \cite{tang2022seafarers}.
			
			5) Lifestyle: It is evaluated by Health-Promoting Lifestyle Profile-II (HPLP) \cite{teng2010health}.
			
			6) ES: It is evaluated by multidimensional fatigue inventory (MFI) \cite{donovan2015systematic}, GAD-7 items and patient health questionnaire 9 (PHQ) \cite{stocker2021patient}, Depression Anxiety and Stress Scale 21 (DASS) \cite{baygi2021prevalence}.
			
			7) WE: It includes company culture, equipment management and maintenance, office environment, safety.
			
			8) En: It includes type of entertainment, frequency of participation in activities.
			
			9) AL: It is evaluated by suicide behaviors questionnaire-revised (SBQ-R) \cite{huen2022suicidal}.

			10) SS: It is evaluated by social support rating scale( SSRS) \cite{tang2022seafarers}.
			
			11) FR: It is evaluated by family assessment device-general functioning (FAD-GF) \cite{tang2022seafarers}. 
			\\\bottomrule
		\end{tabularx}%
	}
\end{table*}%

\begin{figure}[htbp]
	%\begin{center}
	\includegraphics[width = 9.3 cm]{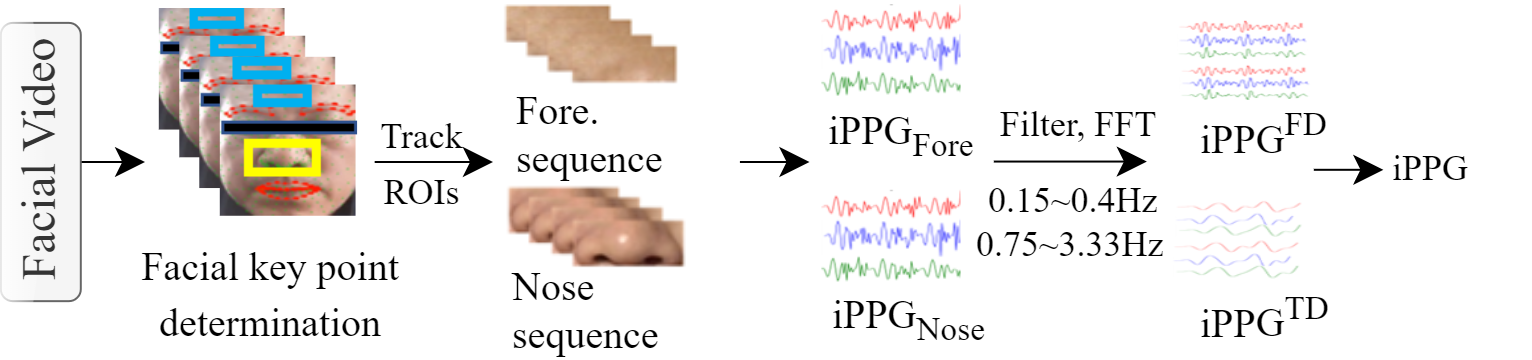}
	%\end{center}
	\caption{Heart rate and respiration rate features extraction.}
	\label{Fig1}
\end{figure}

Patients with anxiety disorders experience overt clinical symptoms such as muscle tension, a racing heartbeat, rapid breathing \cite{efinger2019distraction}, elevated Blood Pressure (BP) \cite{pham2021heart}, and dizziness. One of the key aspects for identifying anxiety is the iPPG signals, which include Heart Rate (HR) and Respiratory Rate (RR). The face's motion-insensitive regions \cite{mo2022collaborative}, such as the forehead and nose, can be used for extracting iPPG characteristics that contain information on HR and RR \cite{ding2022noncontact}. Therefore, the time-domain and frequency-domain features from iPPG signals in the range of HR and RR are used as one of the features for anxiety inference.

Figure 2 depicts the process of extracting HR and RR features from iPPG signals. Each facial video is used for extracting pictures frame by frame. The key feature points from the face are extracted from each frame using a face detection algorithm, such as Google media pipe. The Region of Interest (ROI) from the face in each frame of the picture, such as the forehead and nose, can be precisely tracked based on the key feature points. The ROI (such as $ROI_{Fore}$ and $ROI_{Nose}$) in each frame is resized to a fixed length and width. The ROI's various channels' pixel averages for each frame are used to create iPPG signals. Equation (1) is used to calculate the ROI's Mean Pixel (MP) value for the red channel in the \textit{t}-th frame. In Equation (2), the MP of all frames constitutes the initial iPPG signal.
\begin{equation}
	MP_R(t)=\frac{1}{H_{R O I} \times W_{R O I}} \sum_{x=1}^{H_{R O I}} \sum_{y=1}^{W_{R O I}} P_R(x, t, y)
\end{equation}
\begin{equation}
	iPPG=\left[\begin{array}{l}
		M P_R(1), M P_R(2), \ldots, M P_R(t) \\
		M P_G(1), M P_G(2), \ldots, M P_G(t) \\
		M P_B(1), M P_B(2), \ldots, M P_B(t)
	\end{array}\right]_{C \times T^{\prime}}
\end{equation}
where the red channel pixel value at position (\textit{x, y}) in the \textit{t}-th frame is represented by $P_R$(\textit{x, t, y}). Similarly, the green and blue channel pixel values at position (\textit{x, y}) in the \textit{t}-th frame are denoted as $P_G$ (\textit{x, t, y}) and $P_B$ (\textit{x, t, y}), respectively. The number of channels in each picture frame and the total number of frames in the video are denoted by $C$ and $T'$, respectively. The signals from the nose and forehead are denoted as $iPPG_{Nose}$ and $iPPG_{Fore}$, respectively.

In addition, the initial iPPG signals are processed by Butterworth filter \cite{mo2022collaborative} and Fast Fourier Transform (FFT) to obtain time domain and frequency domain signals with heartbeat and respiration information. The normal heartbeat and breathing of humans are in the range of [0.75, 3.33] Hz, and [0.15, 0.40] Hz, respectively. The time-domain signals within the normal HR and RR range of human body are separated from the initial iPPG signal by filtering, which are called $iPPG_{Fore}^{TD}$ and $iPPG_{Nose}^{TD}$, respectively. The frequency domain signals with HR or RR are extracted from the iPPG signals using the FFT. FFT is able to extract frequency domain features (such as $iPPG_{Fore}^{FD}$ and $iPPG_{Nose}^{FD}$) from HR or RR that as well as efficiently minimizing the noise of the iPPG signals. The iPPG features $iPPG=[iPP{G^{TD}}, iPP{G^{FD}}]$ from the time and frequency domains in the iPPG signals are denoted as $iPP{G^{TD}} = {[iPPG_{Fore}^{TD},iPPG_{Nose}^{TD}]_{2 \times C \times T'}}$ and $iPP{G^{FD}} = {[iPPG_{Fore}^{FD},iPPG_{Nose}^{FD}]_{2 \times C \times T'}}$, respectively.

\subsubsection{Behavioral Features}

Key signs of anxiety in the behavioral context include changes in pupil size, blink rate \cite{zhang2020fusing}, gaze distribution \cite{mogg2007anxiety}, lip twisting, mouth shape movements, cheek changes, head movement, and head speed \cite{adams2015decoupling}. These behavioral features can be described by facial Action Units (AUs) and used as characteristics for anxiety inference.

The different AUs of the face are defined by the Facial Action Coding System (FACS) \cite{liu2019facial}, which describes muscle movements in particular locations. They are used to describe a person's facial activity, including those of the mouth, chin, lips, eyes, eyelids, and eyebrows. Different combinations of AUs can be used to describe various facial behaviors or emotional expressions. A facial behavior analysis toolkit  \cite{baltrusaitis2018openface} is used for analyzing the behavioral features of each frame from the facial video.  $EG(t) = ({x_{EG}}(t),{y_{EG}}(t))$, $AUs(t)$, $HP(t) =({x_{HP}}(t),{y_{HP}}(t),{z_{HP}}(t))$ are Eye Gaze (EG), AUs, and Head Posture (HP) features extracted from the \textit{t}-th frame, respectively. Each frame in the video is used to extract behavioral features, and they are combined to create a sequence of behavioral features $BF$ = [$EG$, $AUs$, $HP$].
\begin{equation}
	EG=[EG(1),EG(2),...,EG(t)]
\end{equation}
\begin{equation}
	AUs=[AUs(1),AUs(2),...,AUs(t)]
\end{equation}
\begin{equation}
	HP=[HP(1),HP(2),...,HP(t)]
\end{equation}

\begin{figure}[htbp]
	%\begin{center}
	\includegraphics[width = 8.5 cm]{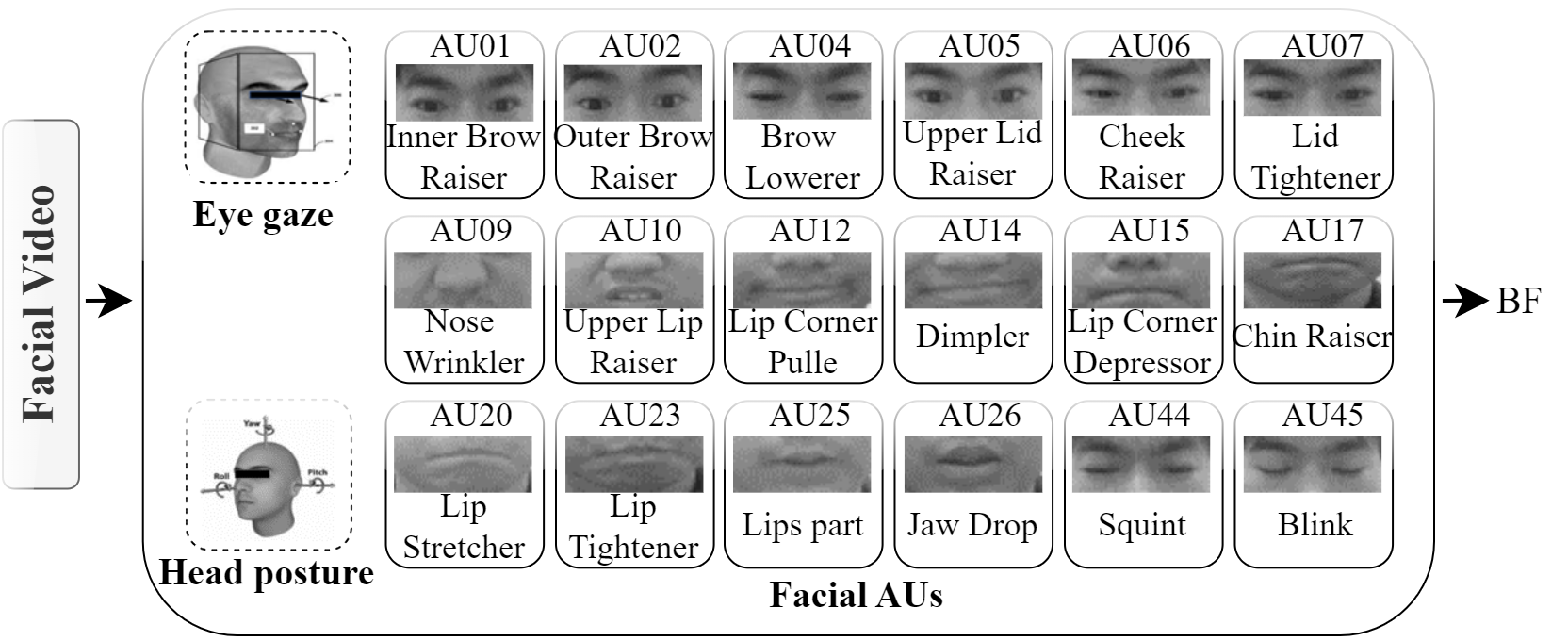}
	%\end{center}
	\caption{Behavioral feature extraction.}
	\label{Fig2}
\end{figure}

\subsubsection{Audio Features}

Negative emotions, such as anxiety, and depression, may cause changes in the somatic and autonomic nervous systems that are reflected in muscle tension and respiratory system \cite{mcginnis2019giving}. These changes can have an impact on prosody and speech quality. In \cite{mcginnis2019giving}, the mean, median, standard deviation, maximum, and minimum of the time domain and frequency domain signals of Zero-Crossing Rate (ZCR) in each sliding time window are extracted. Previous research has demonstrated a correlation between a few auditory factors and anxiety, including fundamental frequency F0 \cite{albuquerque2021association}, the first and second formant frequency (F1 and F2), phonation, MFCCs, and wavelet coefficients. The Phonation consists of the F0's first and second derivatives  ($F0^1$ and $F0^2$), as well as jitter, shimmer, Amplitude Perturbation Quotient (APQ), Pitch Perturbation Quotient (PPQ) and Logarithmic Energy (LE). Partial acoustic signatures vary in different directions and intensities with anxiety. Anxious seafarers have higher F0 mean, F1 mean, jitter, shimmer, and wavelet coefficients. MFCCs are decreased with anxiety \cite{ozseven2018voice}. As a result, these audio features as shown in Table 2 are also used as one of the characteristics for anxiety inference.

There may have background noises in the original audio due to the background environment. It is necessary to eliminate the background noises from the audio and extract the seafarer's voice from each video. Only seafarers' voices are present in the audio data after vocal separation. Initial audio features, such as MFCCs, F0, ZCR, prosody, and phonation, are obtained from the audio data after they have been processed by audio analysis technology \cite{pariente2020asteroid}. Phonation features include $F0^1$, $F0^2$, jitter, shimmer, perturbation quotient, pitch perturbation quotient, and logarithmic energy. Finally, MFCCs, F0, Phonation, Prosody and ZCR form the audio features $AF$=[$MFCCs$, $F0$, $Phonation$, $Prosody$, $ZCR$].

\subsubsection{Text Features}

\begin{figure}[htbp]
	%\begin{center}
	\includegraphics[width = 8.5 cm]{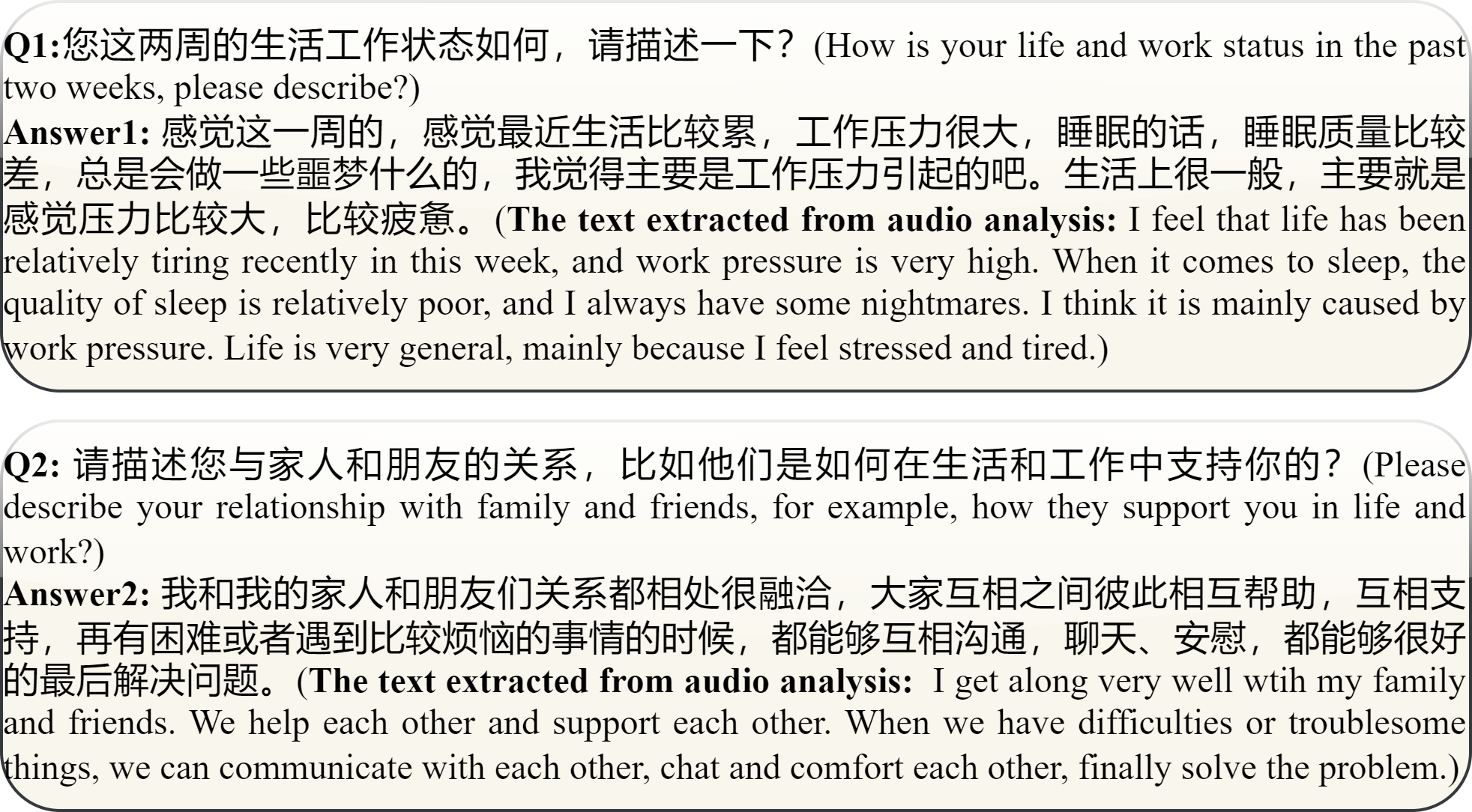}
	%\end{center}
	\caption{Extracted text features based on the answers from seafarers' relationship with relatives and friends, and work status.}
	\label{Fig3}
\end{figure}

When a person works in a closed environment for a long time, his relationship with relatives and friends, and work status are closely related to anxiety \cite{tang2022seafarers}. By asking these questions shown in Figure 4, seafarers' phone cameras record audio and facial video of their responses to learn more about their health. Therefore, text features from audio are also used as the main features for anxiety inference.

The main steps of text feature extraction from audio are as follows. First, human voice is obtained after it has undergone vocal separation processing. Next, the iFlytek toolkit \cite{liu2022spontaneous} with 97.5\% accuracy is used for Chinese voice analysis, and then the Chinese text is extracted from the audio. Finally, the pre-trained BERT model \cite{devlin2018bert} is used to process each sentence from the text in the audio to create the text feature vector. Figure 4 shows a sample of the seafarer's responses to the two questions. Seafarers answer questions based on their current situation.

\subsubsection{Questionnaires Features}

As a result of turbulence, an airtight environment, vibration noise, variations in circadian rhythm, a monotonous diet, and social isolation, long-haul seafarers are exposed to serious health risks \cite{tang2022seafarers}. It can easily trigger a variety of physical and mental health problems for seafarers, including those related to anxiety, diet, illness, fatigue, depression, and cognition. Many studies have found that a variety of factors such as personality traits \cite{nikvcevic2021modelling}, poor sleep quality (leading to fatigue) \cite{tang2022seafarers}, bad emotional state, attitude to life \cite{huen2022suicidal}, a lack of family and social support \cite{brooks2022mental}, and so on, can contribute to anxiety.

In the questionnaire, most of the questions in Table 2 provide options to be accepted which represent different severity or grade levels. Table 3 shows some example questions in the questionnaire. Therefore, each question in the questionnaire can be assigned with a score. The questionnaire features are processed and then denoted as QF.

\begin{table}[]
	\caption{Sample questions in the questionnaire.}
	\scalebox{0.95}{	
		\begin{tabular}{@{}llllll@{}}
			\toprule
			Questions & \multicolumn{5}{c}{Grade levels: low$\to$high} \\ 
			\midrule
			1) I feel energized. & 0  & 1  & 2  & 3  &\CheckedBox 4 \\ 
			2) Have you ever thought about suicide? & 0  & 1  & 2  & 3  & \CheckedBox 4 \\
			3) The working environment is comfortable. & 0  & 1  & 2  & 3  &\CheckedBox 4\\
			4) Difficulty falling asleep. & 0  & 1  & 2  &  3  & \CheckedBox 4 \\
			5) Limit sugar and sugary foods. & 0  & 1  & 2  &  3  & \CheckedBox 4 \\
			6) I often feel exhausted. & 0  & 1  & 2  &  3  & \CheckedBox 4 \\
			7) It's hard to relax.& 0  & 1  & 2  &  3  & \CheckedBox 4 \\
			8) Family members rarely talk to each other. & 0  & 1  & 2  &  3  & \CheckedBox 4 \\
			9) Your colleagues often care about each other.  & 0  & 1  & 2  &  3  & \CheckedBox 4\\
			10) Often receive support and care from family.  & 0  & 1  & 2  &  3  & \CheckedBox 4\\\bottomrule
		\end{tabular}
	}
\end{table}

\subsection{Dimension Reduction}

The majority of features are time series data. The data dimensions of the original HR and RR features from the time domain and frequency domain, behavioral features, and text features from audio are often rather high. For example, a total of 18 AUs can be extracted from each frame of a facial picture given in Figure 3, so the total number of dimensions of AUs obtained from a video with a sampling rate of 25 Frames Per Second (FPS) and a duration of one second is 18 $\times$ 25. It is similar to the dimensions of other time series features. To reduce the data dimensions and the cost of feature selection in the next step, the original features are processed by deep learning networks \cite{petrescu2020integrating}, \cite{tyshchenko2018depression} for dimension reduction, which is shown in Figure 5.

\begin{figure}[htbp]
	%\begin{center}
	\includegraphics[width = 8.5 cm]{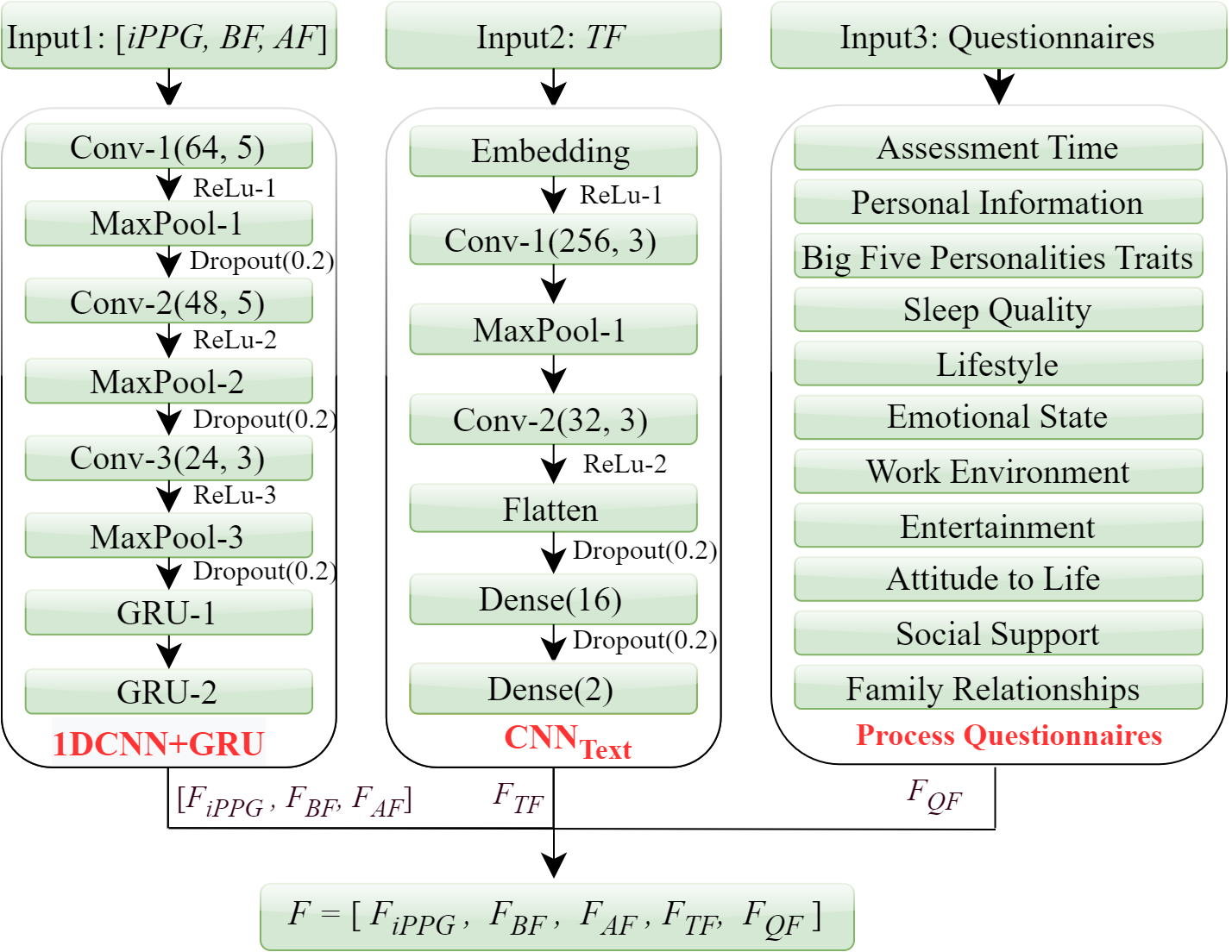}
	%\end{center}
	\caption{Dimension reduction by the "1DCNN+GRU+CNN$_{Text}$".}
	\label{Fig5}
\end{figure}

The Convolutional Neural Network (CNN) can effectively extract spatial features of high-dimensional data \cite{umrani2022hybrid}, while Gated Recurrent Units (GRU) network is a variant of the Long Short-Term Memory (LSTM), which can extract temporal features in time series data \cite{kanai2017preventing}. Compared with time series data, the dimensions of questionnaire features are not high, so it does not need dimension reduction. Therefore, as depicted in Figure 5, the feature vector $[F_{iPPG}, F_{BF}, F_{AF}]$ and $F_{TF}$ are produced when the iPPG, behavior, and audio, and text features are processed by the "1DCNN+GRU" and "CNN$_{Text}$" networks, accordingly. Then, the feature vector $F = [F_{iPPG}, F_{BF}, F_{AF}, F_{TF}, F_{QF}]$ is composed of $[F_{iPPG}, F_{BF}, F_{AF}]$, $F_{TF}$, and $F_{QF}$.

Several "1DCNN+GRU" networks with the same structure are employed to extract spatiotemporal features from iPPG, behavior, and audio respectively to reduce data dimensions. The "1DCNN+GRU" network consists of three 1D-MaxPool layers, three 1D-convolutional layers, and two GRU layers.  A Rectified Linear Unit (ReLU) layer and a dropout layer are added after each convolutional layer and MaxPool layer, respectively. The \textit{Conv-k}(\textit{p1}, \textit{p2}) denotes the \textit{k}-th convolutional layer. The \textit{p1} and \textit{p2} represent the out channels and kernel size, respectively. Additionally, the $TF$ features are processed by the "CNN$_{Text}$" network to obtain the feature vector $F_{TF}$. The "CNN$_{Text}$" network shown in Figure 5 consists of one embedding layer, two convolutional layers, two ReLU layers, one pooling layer, one flatten layer, two dropout layers, and two Dense layers.

In Figure 5, convolutional layers are used to extract different features of the input. The first layer of convolutional layers may only be able to extract some low-level features, and more layers of networks can iteratively extract more complex features from low-level features. The max-pooling layer not only reduces the feature dimension but also retains more texture information. The ReLU layer is to increase the nonlinear fitting ability of the neural network. The dropout layer can enhance the generalization ability of the model. The embedding layer is to encode sentences or words. The flattened layer is to make the multi-dimensional input one-dimensional, and it is often used in the transition from the convolutional layer to the fully connected layer. The dense layer is to change the previously extracted features through nonlinear changes, extract the association between these features, and finally maps them to the output space.

\subsection{Feature Selection}

There are still a lot of redundant features in $F$=[$F_{iPPG}$, $F_{BF}$, $F_{AF}$, $F_{TF}$, $F_{QF}$] extracted from the original data. In addition, due to the data imbalance problem, the parameter learning of the model is more biased toward the majority classes during the training process. As such, the performance of the model will be adversely affected. The feature selection \cite{vsalkevicius2019anxiety} approaches, such as the filter and wrapper methods, are used for anxiety screening to remove redundant features from the original data to enhance the effectiveness of models. Embedding methods, which combine the advantages of filter and wrapper techniques, can also be used for anxiety screening. Moreover, bio-inspired methods that introduce randomness into the search process to avoid local optimum to learn the model parameters are more conducive for predicting the minority class. Therefore, the feature selection method based on our Improved Fireworks Algorithm (IFA) is used to search for feature subsets to solve the feature combination optimization problem, which is non-differentiable.

\subsubsection{Improved Fireworks Algorithm}

Fireworks Algorithm (FA) \cite{li2019comprehensive} is a new swarm intelligence algorithm proposed in recent years, and its idea comes from the fireworks explosion process in real life. FA automatically balances local and global search capabilities by regulating the quantity of offspring generated by fireworks through the explosion intensity. The former can hasten population convergence, whilst the latter can guarantee population diversity. The original FA uses the explosion, mutation, selection, and mapping rules as its four major operators. Based on the original FA \cite{li2019comprehensive}, our Improved Fireworks Algorithm (IFA) enhances the explosion radius (also called explosion amplitude) of FA in the explosion operator to improve the local search capability by Equations (6) and (7) while leaving the other components of the algorithm unaltered.

\begin{equation}
	R_i^{\text {new }}=\left\{\begin{array}{l}
		x_{C F} \times(1+N(0,1))-x_i, S_i=S_{\max } \\
		R_{\max } \frac{f\left(x_i^{\text {pbest }}\right)-Y_{\min }^{\text {pbest }}+\varepsilon}{\sum_{i=1}^N\left(f\left(x_i^{pbest}\right)-Y_{\min }^{\text {pbest }}\right)+\varepsilon}, S_i \neq S_{\max}
	\end{array}\right.
\end{equation}
\begin{equation}
	x_i^{pbest}=\left\{\begin{array}{l}
		x_i, f\left(x_i\right)<f\left(x_i^{pbest}\right) \\
		x_i^{pbest}, \text { \textit{otherwise} }
	\end{array}\right.
\end{equation}
where the Core Fireworks (CF) $x_{CF}$ is the individual with the best fitness value in the fireworks population. The Gaussian distribution function $N(0, 1)$ has a mean of zero and a variance of 1. $S_{i}$ is the number of explosion sparks produced by the \textit{i}-th firework individuals $x_{i}$. $S_{max}$ is the maximum number of explosion sparks. $R_{max}$ is the maximum explosion radius that the fireworks individuals are allowed to displace. The \textit{i}-th firework individual’s fitness value is represented by $f(x_i)$. The worst firework individual’s fitness value is $Y_{max}$ = max\{$f(x_1)$, $f(x_2)$, ..., $f(x_i)$\}.

The FA chooses the next generation of fireworks from candidate individuals, and the others are discarded. The best historical information in the candidate set will not be fully utilized using this strategy in the FA. Equations (6) and (7) illustrate how our IFA generates an adaptable explosion radius by utilizing the historical information $x_{i}^{pbest}$  from the \textit{i}-th fireworks individual $x_{i}$. If the \textit{i}-th fitness value of $x_i$ is smaller than that of  $x_i^{pbest}$, the $x_i^{pbest}$ is updated by Equation (7).

\subsubsection{IFA-based Feature Selection}

\begin{figure}[htbp]
	%\begin{center}
	\includegraphics[width = 8.5 cm]{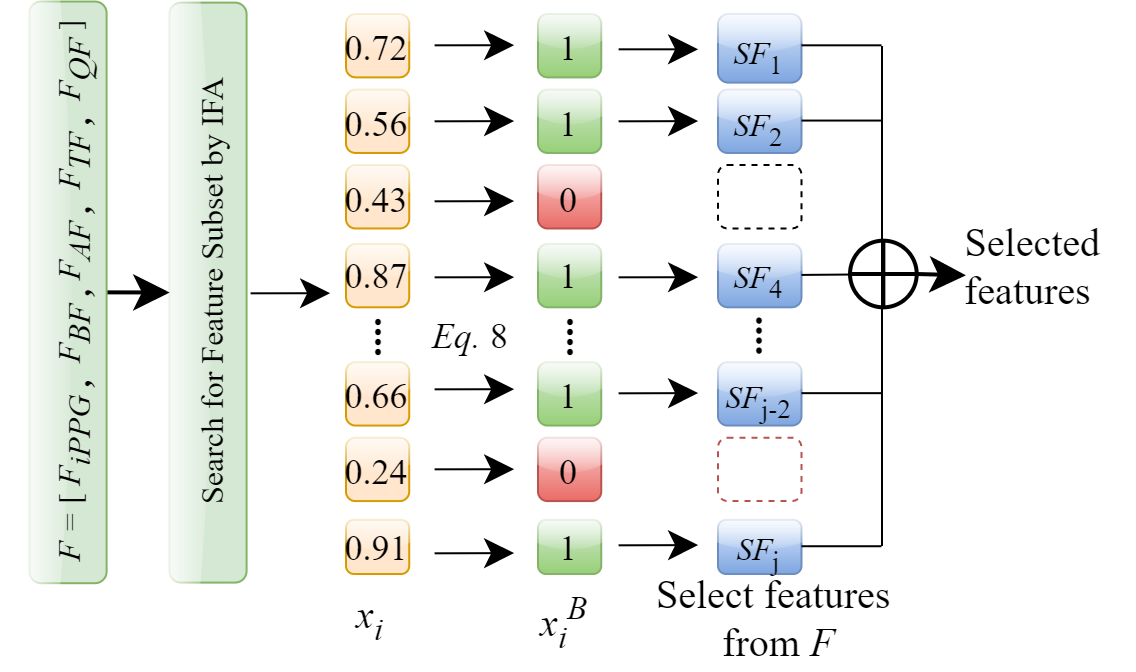}
	%\end{center}
	\caption{Feature selection based on the Improved Fireworks Algorithm.}
	\label{Fig5}
\end{figure}

The process of feature selection based on the Improved Fireworks Algorithm is shown in Figure 6. The purpose of the IFA iteration is to search for the feature subset's locations. In other words, each individual $x_{i}$ of the IFA can represent a set of feature subsets, which are used to determine the corresponding dimension of the selected features from the feature vector $F=[F_{iPPG}, F_{BF}, F_{AF}, F_{TF}, F_{QF}]$.

For example, the value of each dimension of individual $x_{i}$ generated by each iteration of IFA is in the range of [0,1]. After $x_{i}$ is discretized by Equation (8), it can represent a set of decision variables. 

\begin{equation}
	x_{ij}^B = \left\{ \begin{array}{l}
		0, {x_{ij}} < 0.5\\
		1, otherwise
	\end{array} \right.
\end{equation}
where $x_{ij}^B$ represents the \textit{Binary} value of the \textit{j}-th dimension of the \textit{i}-th individual $x_{i}$ after being discretized. That is to say, when $x_{ij}<0.5$ and $x_{ij}^B=0$, it means that the feature corresponding to the \textit{j}-th dimension in the feature vector $F$ is not selected, otherwise it is selected. The green squares with the value in Figure 6 represent the index of the selected feature, while the red squares represent the index of the unselected feature at the discretized position of the \textit{i}-th individual $x_{i}^B$. Finally, the Selected Features $SF=[SF_{1}, SF_{2}, SF_{4}, ..., SF_{j-2}, SF_{j}]$ can be determined from feature vector \textit{F} by $x_{i}^B=[x_{i1}^B, x_{i2}^B,...,x_{ij}^B]$, and $j$=1, 2, ..., $d$. And $d$ is the total number of dimensions of $F$.

\subsection{Anxiety Inference}

\begin{figure}[htbp]
	\includegraphics[width = 8 cm]{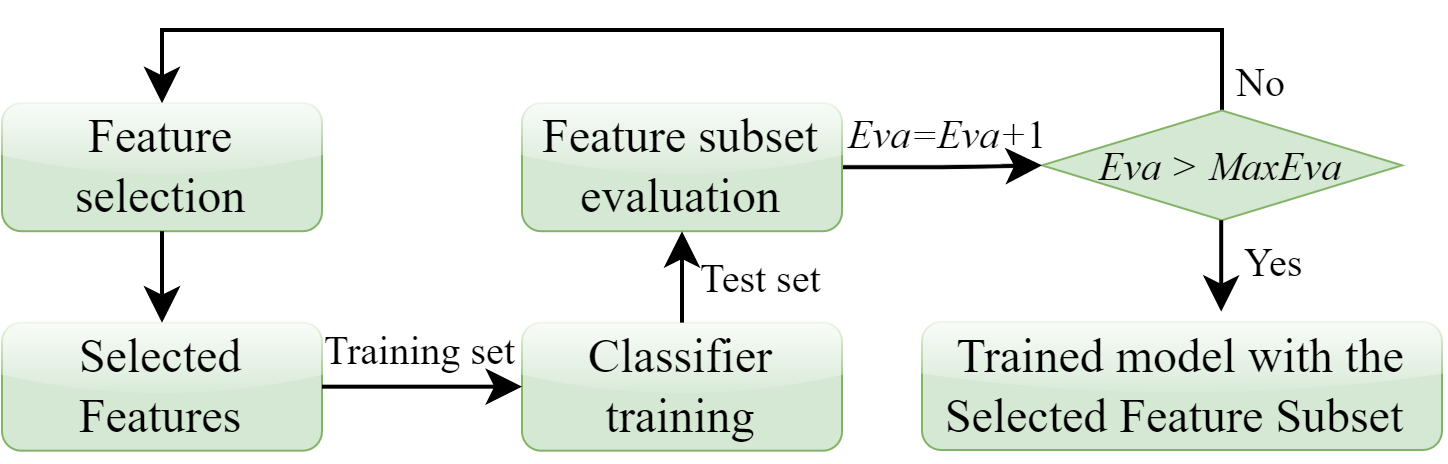}
	\caption{Anxiety Inference.}
	\label{Fig5}
\end{figure}

The parameter learning for anxiety inference models can be viewed as a 0-1 programming problem. If the feature vector $F$ in Figure 6 has \textit{d}-dimensional features, there are $2^{d}$ possible feature combinations. The feature subset is searched by IFA from the solution space of $2^{d}$ feature combinations to find the one that can achieve the best model's overall classification evaluation metrics. In previous studies \cite{zhang2020fusing}, the model's performance usually uses one or more classification evaluation metrics, such as accuracy, precision, sensitivity, specificity, or F1-score. However, when the samples in the data set are unbalanced, these metrics can hardly distinguish the model's performance \cite{mullick2020appropriateness}. In addition, too low sensitivity or specificity may cause adverse consequences. For instance, a test with low sensitivity may cause errors, and fail to detect correctly. The test's low specificity might lead to a lot of false positive results, which is quite stressful for patients. Therefore, it is necessary to measure the performance of the model by multiple classification evaluation metrics, such as the Area Under Curve (AUC), accuracy (Acc), precision (Pre), sensitivity (Sen), specificity (Spe), and F1-score (F1), and use Equation (9) as the loss function for model optimization. A penalty factor $\lambda=0.2 \times d$ needs to be introduced to limit the dimensions of the selected features.
\begin{equation} 
	\begin{split} 
		minf(x_i^{B}) =- (AUC + Acc + Pre+ Sen +\\ F1 + Spe), 
		\sum\limits_{j = 1}^d {x_{ij}^B}  \ge \lambda 
	\end{split} 
\end{equation}
\begin{equation}
	Acc = \frac{{TP+TN}}{{TP + TN + FP+FN}}
\end{equation}
\begin{equation}
	Pre = \frac{{TP}}{{TP + FP}}
\end{equation}
\begin{equation}
	Sen = \frac{{TP}}{{TP + FN}}
\end{equation}
\begin{equation}
	F1 = \frac{{2TP}}{{2TP+FP+FN}}
\end{equation}
\begin{equation}
	Spe = \frac{{TN}}{{TN + FP}}
\end{equation}
where True-Positive (TP) and True-Negative (TN) denote the accurate classification of the anxiety and anxiety-free samples, respectively. False-Negative (FN) and False-Positive (FP) suggest that the anxiety and anxiety-free samples are wrongly categorized respectively. 
In addition, AdaBoost \cite{zhao2019adaptive} is a technique for ensemble learning that uses an iterative process to improve weak classifiers by learning from their mistakes \cite{giannakakis2017stress}, \cite{puli2019toward}. Due to its effective classification performance and the interpretability of the inference process, Adaboost is utilized as the classifier for anxiety screening.

Figure 7 depicts the anxiety inference process. The feature subset produced by the feature selection method can be used for determining the selected features. The data corresponding to these features is divided into training and test sets. The training set is used to train the classifier. The trained classifier assesses the feature subset based on the predictions from the test set. When a feature subset is evaluated once, the number of evaluations is increased accordingly, ie, $Eve = Eve + 1$. If the total number of evaluations reaches the predefined maximum number of evaluations, ie, $Eve > MaxEve$, the selected features and trained classifier are used for anxiety inference. Otherwise, the feature selection method process searches for a new feature subset.

\section{Performance Evaluation}

To gather information about the physical and mental health from seafarers for conducting the experiments, we have collaborated with West China Hospital of Sichuan University for performance evaluation. These experiments mainly focus on seafarers' health perception and intervention. On August 12, 2020, the Biomedical Ethics Committee of Hefei University of Technology gave the study its full approval with experiment registration number W2020JSFW0388. All participating seafarers had given the consent to the experiments.

\subsection{Experiments}

\begin{figure}[htbp]
	%\begin{center}
	\includegraphics[width = 8.5 cm]{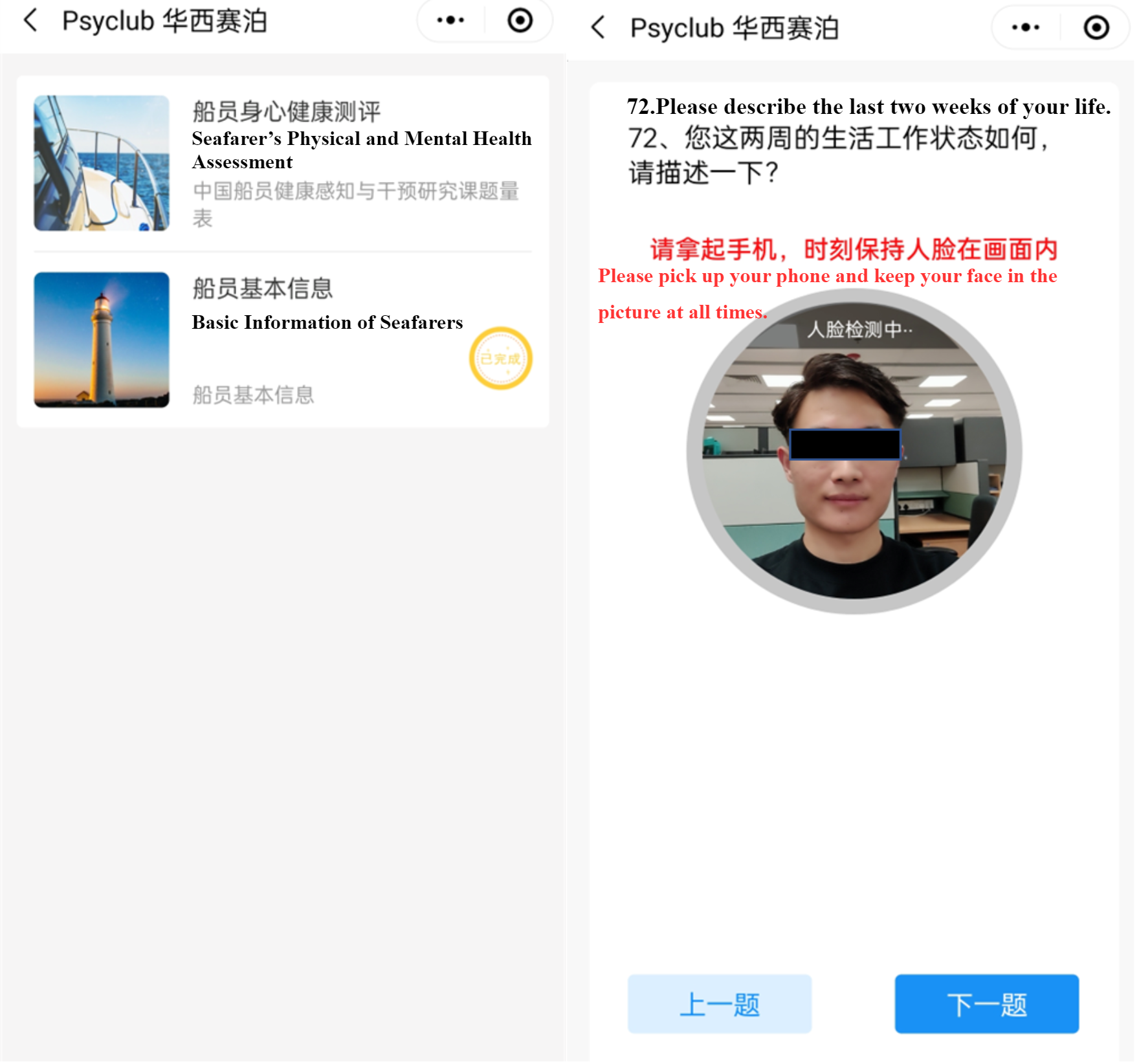}
	%\end{center}
	\caption{The interface of seafarers' physical and mental health assessment.}
	\label{Fig7}
\end{figure}

We have designed a system that can be accessed using mobile devices based on the WeChat applet platform. With their smartphones, seafarers can then check their physical and mental well-being. First, seafarers need to fill in a questionnaire. As shown in Table 2, the content of the questionnaires includes personality traits \cite{nikvcevic2021modelling}, poor sleep quality (leading to fatigue) \cite{tang2022seafarers}, bad emotional state, attitude to life \cite{huen2022suicidal}, family relationship, social support \cite{brooks2022mental}, etc.

Next, the seafarers are then asked additional questions on their work status, and relationships with family and friends after completing the questionnaires. The seafarer's responses are captured on camera by their phones. Each video capture lasts for 30 seconds. The smartphone concurrently records the audio data while recording the video. The sampling rates for audio and video are 25 FPS and 22.05 kHz, respectively. The video has a 480 $\times$ 480 pixel resolution. The average, maximum, and minimum audio durations are 29.31 seconds, 30.44 seconds, and 28.96 seconds, respectively. When responding to the inquiries, the seafarer's face is kept as visible as possible on the recorded screen, as shown on the right side of the interface in Figure 8. One frame per second is taken to assess the acquired video’s quality before the video is transferred to the server. The video is uploaded to the server if the rate of faces being detected in it exceeds 90\%. Otherwise, it needs to be captured again. 

In the experiments, all seafarers taking part in the physical and mental health evaluation were male. The age range of seafarers was 19 to 58. There were 2 seafarers' age between 18 and 20, 167 people between the ages of 20 and 40, and 20 seafarers' age between 40 and 60. The seafarers’ collective age was known to be 31, on average. The age information for other seafarers was not recorded due to errors. The start and end dates for data collection were from June 2020 to June 2021. A total of 227 of seafarers' data were recorded: 189 of them had no anxiety, 33 had mild anxiety, and 5 had moderate anxiety. In other words, 189 people were labeled "anxiety-free" and the remaining 38 people were labeled "anxiety". The GAD-7  \cite{stocker2021patient} was used to measure seafarers' anxiety levels. Additionally, psychiatrists were invited to validate the outcomes of the GAD-7 scale that the seafarers had completed.

\subsection{Performance Results}

% Table generated by Excel2LaTeX from sheet 'Sheet3' htbp
\begin{table*}[htbp]
	\centering
	\caption{Performance comparison (\%) of different methods.}
	\scalebox{1}{	
		\begin{tabular}{llllll}
			\toprule
			\multirow{2}[4]{*}{Methods} & \multicolumn{3}{c}{Different components in the methods} & \multirow{2}[4]{*}{Avg(\%)} & \multirow{2}[4]{*}{$\Delta$Avg(\%)} \\
			\cmidrule{2-4}          & Dimension Reduction & Feature Selection & Anxiety Inference  &       &  \\
			\midrule
			\textbf{MMD-AS (ours)} & 1DCNN+GRU+CNN$_{Text}$ & IFA   & AdaBoost & \textbf{97.55}  & - - - \\ 	\midrule
			M1    & 1DCNN+LSTM+CNN$_{Text}$ & IFA   & AdaBoost & 96.71  & -0.84  \\
			M2    & LSTM+CNN$_{Text}$ & IFA   & AdaBoost & 96.27  & -1.28  \\
			M3    & 1DCNN+CNN$_{Text}$ & IFA   & AdaBoost & 97.01  & -0.54  \\
			M4    & PCA   & IFA   & AdaBoost & 58.27  & -39.28  \\
			\midrule
			M5    & 1DCNN+GRU+CNN$_{Text}$ & FA    & AdaBoost & 97.10  & -0.45  \\
			M6    & 1DCNN+GRU+CNN$_{Text}$ & BA    & AdaBoost & \underline{97.36}  & -0.19  \\
			M7    & 1DCNN+GRU+CNN$_{Text}$ & PSO   & AdaBoost & 92.23  & -5.32  \\
			M8    & 1DCNN+GRU+CNN$_{Text}$ & SKB   & AdaBoost & 95.30  & -2.25  \\
			\midrule
			M9    & 1DCNN+GRU+CNN$_{Text}$ & IFA   & DT    & 94.83  & -2.72  \\
			M10   & 1DCNN+GRU+CNN$_{Text}$ & IFA   & RF    & 96.75  & -0.80  \\
			M11   & 1DCNN+GRU+CNN$_{Text}$ & IFA   & LR    & 94.08  & -3.47  \\
			M12   & 1DCNN+GRU+CNN$_{Text}$ & IFA   & KNN   & 80.03  & -17.52  \\
			M13   & 1DCNN+GRU+CNN$_{Text}$ & IFA   & SVM   & 92.43  & -5.12  \\
			M14   & 1DCNN+GRU+CNN$_{Text}$ & IFA   & MLP   & 95.05  & -2.50  \\
			\bottomrule
		\end{tabular}%
	}
	\label{tab:addlabel}%
\end{table*}%

Our proposed MMD-AS framework consists of the dimension reduction component "1DCNN+GRU+CNN$_{Text}$", feature selection component "IFA", and anxiety inference component "AdaBoost". Table 4 shows the performance results of different methods for different components. The performance is computed using Average (Avg) = (Acc + AUC + Pre + Sen + F1 + Spe)/6. $\Delta$Avg shows the average performance difference of the method when compared to the proposed MMD-AS framework. Overall, the proposed MMD-AS has achieved the best performance with Avg = 97.55\%.

\subsubsection{Performance on Dimension Reduction Methods}
The dimension reduction method of the proposed MMD-AS framework is "1DCNN+GRU+CNN$_{Text}$". The methods, which are used for comparison with the framework's dimension reduction method, include "1DCNN+ LSTM+CNN$_{Text}$" (M1), "LSTM+CNN$_{Text}$" (M2) \cite{richards2011influence}, "1DCNN+ CNN$_{Text}$" (M3), and Principal Component Analysis (PCA) \cite{haritha2017automating} (M4). The experiments are conducted by using the same feature selection component "IFA", and anxiety inference component "AdaBoost" \cite{giannakakis2017stress}, \cite{puli2019toward} of the proposed MMD-AS framework with different dimension reduction methods. The performance results of MMD-AS with the dimension reduction method "1DCNN+GRU+CNN$_{Text}$" has achieved the best performance. The proposed MMD-AS framework has achieved performance improvements of 0.84\%, 1.28\%, 0.54\%, and 39.28\%, respectively when compared with M1 to M4. However, the performance of M4 with dimension reduction method PCA performs worse than the methods based on deep learning-based dimension reduction methods such as "1DCNN+GRU+CNN$_{Text}$" (MMD-AS), "LSTM+CNN$_{Text}$" (M2) and "1DCNN+CNN$_{Text}$" (M3). The reason for the performance differences between them is that the dimension reduction methods in MMD-AS and M1 to M3 are deep learning networks, which can extract time-series features from high dimensions and can effectively reduce the original data's dimensions to improve the performance.

\subsubsection{Performance on Feature Selection Methods}

The feature selection method of the proposed MMD-AS framework is IFA. The methods, which are used for comparison with the framework's feature selection method, include Fireworks Algorithm (FA) \cite{li2019comprehensive} (M5), Bat Algorithm (BA) \cite{yang2012bat} (M6), Particle Swarm Optimization (PSO) \cite{praveen2022pso} (M7), and Selecting \textit{K}-Best (SKB)\cite{akman2023k} (M8). The experiments are conducted by using the same dimension reduction component "1DCNN+GRU+CNN$_{Text}$", and the anxiety inference component "AdaBoost" of the proposed MMD-AS framework with different feature selection methods. The proposed MMD-AS framework with the feature selection method "IFA" has achieved the best performance. The proposed MMD-AS framework has achieved performance improvements of 0.45\%, 0.19\%, 5.32\%, and 2.25\%, respectively when compared with M5 to M8.

The proposed MMD-AS framework with the feature selection method IFA has the improved explosion radius, which offers better local search capability to guide the fireworks population to find the better feature subset and reduce the noises in the features. Therefore, The MMD-AS's IFA algorithm outperforms the Fireworks Algorithm (M5) by 0.44\%. However, PSO (M7) and SKB (M8) perform quite poorly when compared with other feature selection methods based on swarm intelligence algorithms such as IFA, FA, and BA. The main reason for the poor performance of PSO (M7) is that the PSO's search capability is not as good as other swarm intelligence algorithms, such as IFA, FA, and BA. In addition, as features with small variance may contain important information that distinguishes samples, the dimension reduction method of SKB may cause these features to be filtered, which may be the reason for the poor performance of SKB (M8).

\subsubsection{Performance on Anxiety Inference Methods}

The anxiety inference component of the proposed MMD-AS framework is AdaBoost. The methods, which are used for comparison with the framework's anxiety inference method, include Decision Tree (DT) \cite{ihmig2020line} (M9), Random Forest (RF) \cite{tyshchenko2018depression} (M10), Logistic Regression (LR) \cite{giannakakis2017stress} (M11), K-Nearest Neighbors (KNN) \cite{li2019recognition} (M12), Support Vector Machines (SVM)\cite{tyshchenko2018depression} (M13), and Multilayer Perceptron (MLP) \cite{alnuaim2022human} (M14).

The experiments are conducted by using the same dimension reduction component "1DCNN+GRU+CNN$_{Text}$", and feature selection component IFA of the proposed MMD-AS framework with different anxiety inference methods. The proposed MMD-AS framework with the anxiety inference method "AdaBoost" has achieved the best performance, with performance improvements of 2.72\%, 0.80\%, 3.47\%, 17.52\%, 5.12\%, and 2.50\% respectively when compared with M9 to M14. Since AdaBoost is an ensemble learning technique that uses an iterative process to improve weak classifiers \cite{giannakakis2017stress}, \cite{puli2019toward} by learning from mistakes, it can learn features that are conducive to anxiety inference.

\subsection{Ablation Study}

% Table generated by Excel2LaTeX from sheet 'V7'
\begin{table*}[htbp]
	\centering
	\caption{Experimental results of the ablation study of the MMD-AS framework.}
	\scalebox{1}{	
		\begin{tabular}{llllll}
			\toprule
			\multirow{2}[4]{*}{Methods} & \multicolumn{3}{c}{Different components in the framework} & \multicolumn{1}{c}{\multirow{2}[4]{*}{Avg(\%)}} & \multicolumn{1}{c}{\multirow{2}[4]{*}{$\Delta$Avg(\%)}} \\
			\cmidrule{2-4}          & Dimension Reduction & Feature Selection & Anxiety Inference  &       &  \\
			\midrule
			\textbf{MMD-AS (ours)} & 1DCNN+GRU+CNN$_{Text}$ & IFA   & AdaBoost & \textbf{97.55 } & - - -  \\
			M15   & 1DCNN+GRU+CNN$_{Text}$ & -     & AdaBoost & 91.90  & -5.65  \\
			M16   & 1DCNN+GRU+CNN$_{Text}$ & -     & -     & 93.99  & -3.56  \\
			M17   & -     & -     & Adaboost & 89.30  & -8.25  \\
			\bottomrule
		\end{tabular}%
	}
	\label{tab:addlabel}%
\end{table*}%

We have conducted an experiment for an ablation study to evaluate the effectiveness of each component in our proposed MMD-AS framework: (1) M15 is MMD-AS without feature selection. (2) M16 is MMD-AS without feature selection and anxiety inference. As M16 only has a dimension reduction component "1DCNN+GRU+CNN$_{Text}$", it cannot perform classification. So we add a fully connected layer behind the component to enable M16 with a classification function. (3) M17 is MMD-AS without dimension reduction and feature selection. 

Table 5 shows the results of the ablation experiments. We have the following observations. First, MMD-AS outperforms M15 by 5.65\%, which demonstrates the capabilities of the feature selection method IFA in terms of feature selection and feature denoising. Second, MMD-AS outperforms M16 by 3.56\%, which shows that the components IFA and AdaBoost improve the model's performance. Third, MMD-AS outperforms M17 by 8.25\%, which shows that the components "1DCNN+GRU+CNN$_{Text}$" and IFA can improve the model's performance. Overall, the different components of the proposed MMD-AS framework are important for the framework to achieve the best performance.

\subsection{Analysis on Feature Selection}

\begin{figure*}[htbp]
	\begin{center}
		\includegraphics[width = 16 cm]{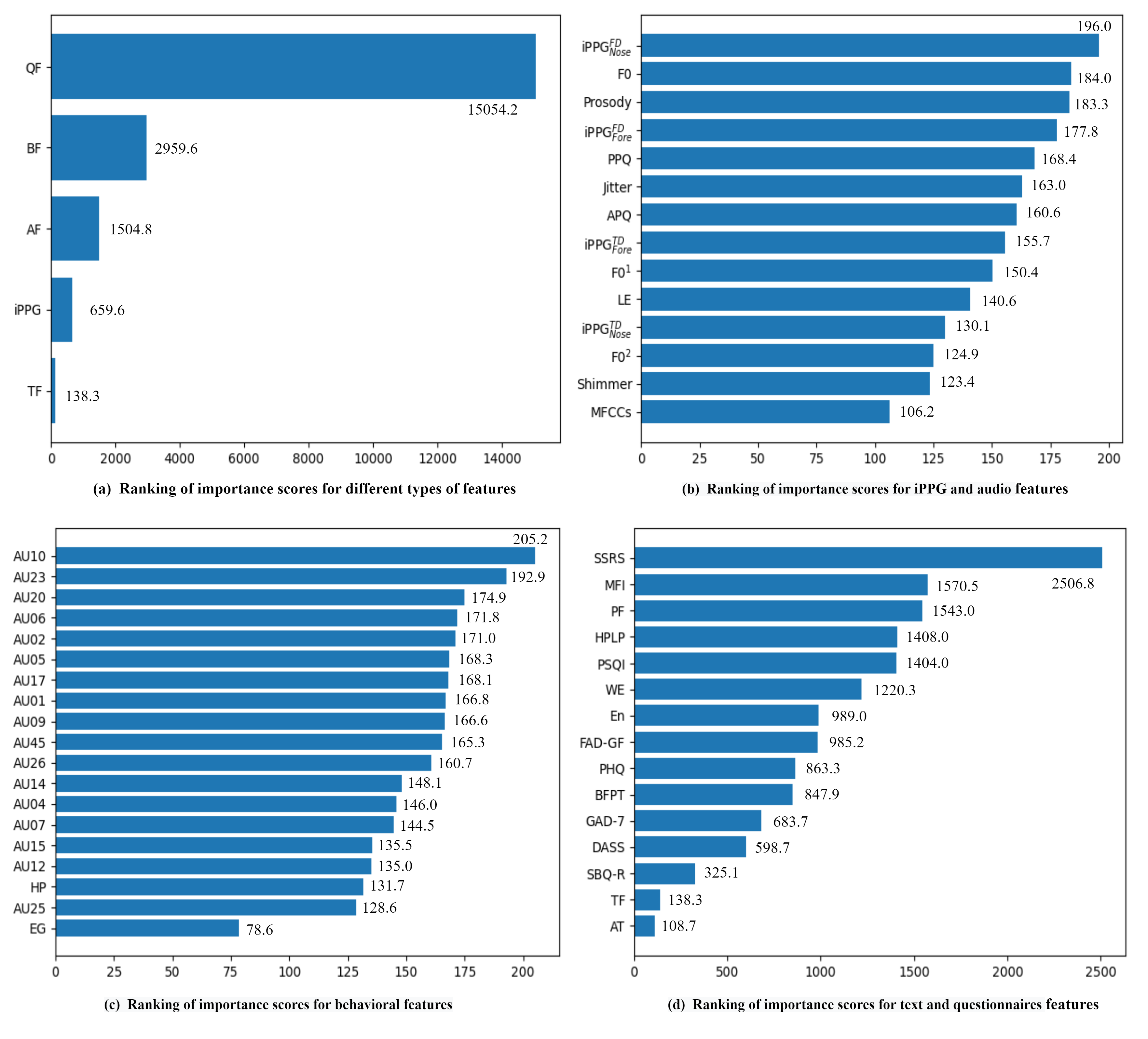}
	\end{center}
	\caption{The importance scores ranking  of the proposed MMD-AS on different features}
	\label{Fig7}
\end{figure*}

Feature importance scores are used to indicate which features are useful for anxiety screening. When the feature selection algorithm selects a feature of a certain dimension in the feature $F$, the importance score of the feature is increased by one. Figures 9(a) to 9(d) show the ranking of our proposed MMD-AS framework on the importance scores of different types of features such as iPPG features (including HR and RR features), BF, AF, TF, and QF for anxiety screening. 

In Figure 9(a), the importance of QF and BF features are ranked the highest and the second highest among the five categories of features, respectively. Figure 9(b) shows the feature importance scores of physiological representations in anxiety screening, such as audio features, and iPPG containing heart rate and respiration rate features. Audio features, such as Prosody, fundamental frequency features (including F0 and F0$^1$), Pitch Perturbation Quotient (PPQ), Jitter, and Amplitude Perturbation Quotient (APQ), play a more important role in anxiety screening. Their feature importance scores are all greater than 150. The iPPG features from frequency domain signals including iPPG$_{Fore}^{FD}$ and iPPG$_{Nose}^{FD}$ are more important for anxiety screening. Compared with the iPPG features from the nose area, iPPG features from the forehead are more important for anxiety screening, since the forehead is dense with blood vessels \cite{mo2022collaborative}. Figure 9(c) demonstrates that anxiety screening is more heavily influenced by characteristics in the chin (e.g. AU17), eyes (e.g. AU02, AU05, AU01, AU45, AU04), and lips (e.g. AU10, AU23, AU20) areas. The scores of these features are mostly distributed in the range of [160, 206]. Figure 9(d) shows the contribution of text and questionnaire features to anxiety screening. The feature importance scores of SSRS, MFI, PF, HPLP, and PSQI features ranked among the top in anxiety screening. The importance scores of these features are all greater than 1400.

Due to the complex etiology and long development cycle of anxiety, it usually requires a combination of multidisciplinary knowledge such as biomedicine, psychology, and social medicine to assist diagnosis \cite{farre2017new}. In clinical practice, multimodal data are essential for anxiety screening \cite{zhang2020fusing}, \cite{lee2020detection}. Therefore, based on the results of the feature analysis, we suggest that screening patients with anxiety should be combined with multimodal information. In addition, behavioral features (such as chin, eyes, lip area), physiological signals from the frequency domain (such as heart rate, respiration rate), audio characteristics (such as the Prosody, fundamental frequency features, PPQ, Jitter, and APQ), social support, fatigue status, sleep quality \cite{tang2022seafarers}, lifestyle and other important modal characteristics can be used as indicators for clinical practice.

\section{Limitations}

There are still some shortcomings with our proposed framework.

First, our dataset has some limitations, such as uneven data and the small sample size. Due to the inherent characteristics of the seafarers' profession, our dataset lacks health data related to female seafarers. Based on the results of health examinations, shipping companies have restricted seafarers with severe mental illness from boarding the ships for work, which may lead to the lack of data samples of severe anxiety in our dataset. Our framework can only learn from the existing samples, and not from the parameters in the missing samples. Our framework is driven by multimodal data, which may result in our model not being able to effectively screen out people suffering from anxiety when faced with certain new samples, which in turn limits the generalization and application.

Second, in the standard clinical diagnostic process, physicians are required to conduct a structured interview with the patient to further determine the patient's mental health status. Although the GAD-7 is used to label seafarers' anxiety levels, the lack of structured interviews with seafarers may lead to misdiagnosis or underdiagnosis of a small number of seafarers. However, multimodal data can provide physicians with more objective evidence when screening for anxiety in seafarers. Psychologists re-examine the psychological scales completed by seafarers, which reduces the probability of misdiagnosis or underdiagnosis of seafarers.

Third, important features can serve as objective evidence for anxiety reasoning, which needs more cohort experiments to verify. Since the factors leading to the formation of anxiety are multifaceted, which involve biological, medical, and sociological aspects. Therefore, it is very necessary to further validate this conclusion by increasing the data sample size and conducting cohort experiments. Nonetheless, these important characteristics are beneficial to design cohort research experiments to investigate the mechanisms of anxiety development.

Because of the above issues, we will focus on the following three areas in our future research.

First, knowledge from a range of fields, such as biomedical, psychological, and sociomedical, needs to be integrated into the proposed framework to enhance interpretability. By utilizing wearable technology and noncontact technologies, we will focus on extracting features from different dimensions, such as physiological features, heart rate variability, changes in facial temperature, distribution of facial temperature, audio features), behavioral features, family social support, sleep quality, and fatigue status. By combining these characteristics with clinical expert knowledge, more scientific evidence useful for anxiety inference can be provided. In addition, additional objective evidence expert domain knowledge will be integrated into our framework to assist primary care physicians in anxiety screening.

Second, the increased data sample size allows for a more comprehensive data distribution, which can enhance the robustness of the anxiety screening framework. The results of the analysis of important features based on the use of anxiety inference are used to build cohort research experiments, which in turn assist physicians in studying how anxiety is developed.

Finally, our framework offers the advantages of low cost, ease of use, noncontact, interpretable and high accuracy. In addition, our framework enables anxiety screening of seafarers by simply analyzing multimodal health data via smartphones. It is invaluable in future telemedicine scenarios. Therefore, our framework will be extended and applied to anxiety detection for large populations in scenarios where medical resources are limited, such as health coverage for seafarers on long voyages, or in remote areas.

\section{Conclusion}

Existing methods for anxiety screening have some drawbacks, such as the inability to solve the non-differentiable problem of feature combinations and the inability to meet the requirement of scenarios with limited medical resources. To overcome these drawbacks, we have proposed a multimodal data-driven framework called MMD-AS for seafarer's anxiety screening in this paper. The experimental results of comparative experiments and ablation studies of different components in the framework show that our proposed framework has achieved the best performance among the comparison methods, and each component of the proposed framework is important for performance improvement. In addition, due to the advantages of low cost, noncontact and convenient operation, the proposed framework and the suggested indications for anxiety screening have certain guiding significance and application value for application scenarios with limited medical resources, such as the health protection of seafarers on long-distance voyages and anxiety screening in remote areas. For further work, we will collect relevant health data from more people, and apply the proposed framework for anxiety screening in clinical practices, which can provide more detailed and scientific evidence for anxiety screening to help study the process of anxiety development.

\ifCLASSOPTIONcompsoc
  \section*{Acknowledgments}
\else
  \section*{Acknowledgment}
\fi
The authors would like to thank the professors Wei Zhang and Yuchen Li from West China Hospital of Sichuan University for the guidance on experimental design and data collection. This work was supported in part by the 2020 Science and Technology Project of the Maritime Safety Administration of the Ministry of Transport of China (No. 0745-2041CCIEC016) and the National Natural Science Foundation of China (No. 91846107, 72293581, 72293580). 
\ifCLASSOPTIONcaptionsoff
  \newpage
\fi



%\bibliographystyle{IEEEtran}

\bibliographystyle{IEEEtran} 	
\bibliography{reference_IEEE}



%\vfill


\end{document}